\pdfoutput=1

\documentclass[11pt]{article}

\usepackage{acl}

\usepackage{times}
\usepackage{latexsym}
\usepackage{graphicx}
\usepackage{caption}
\usepackage{subcaption}
\usepackage[T1]{fontenc}

\usepackage[utf8]{inputenc}

\usepackage{microtype}

\title{Machine Learning Approaches for Principle Prediction in Naturally Occurring Stories}

\author{Md Sultan Al Nahian\thanks{*Denotes equal contribution}  \textsuperscript{1},
Spencer Frazier\footnotemark[1] \textsuperscript{2},
Brent Harrison\textsuperscript{1},
Mark Riedl\textsuperscript{2}\\
\textsuperscript{1}{University of Kentucky}\\
\textsuperscript{2}{Georgia Institute of Technology}\\
sa.nahian@uky.edu,
sf@gatech.edu, 
riedl@cc.gatech.edu,
harrison@cs.uky.edu}

\begin{document}
\maketitle
\begin{abstract}

Value alignment is the task of creating autonomous systems whose values align with those of humans. 
Past work has shown that stories are a potentially rich source of information on human values; however, past work has been limited to considering values in a binary sense. 
In this work, we explore the use of machine learning models for the task of \textit{normative principle prediction} on naturally occurring story data. 
To do this, we extend a dataset that has been previously used to train a binary normative classifier with annotations of moral principles. We then use this dataset to train a variety of machine learning models, evaluate these models and compare their results against humans who were asked to perform the same task. We show that while individual principles can be classified, the ambiguity of what "moral principles" represent, poses a challenge for both human participants and autonomous systems which are faced with the same task.


\end{abstract}

\section{Introduction}

As the capabilities of artificial intelligence systems expand, there has been an increased need to create systems that can safely interact with human operators. 
This is especially true for systems that are given degrees of autonomy. 
One proposed means of creating safe, autonomous systems is to ensure that the system's values align with those of their human counterparts~\cite{soares2014aligning,russell2015research,arnold2017value}. 
This task, often referred to as \textit{value alignment}, is a critical task as it is assumed that this will guard against potential harmful interactions between humans and AI systems~\cite{russell-new-book,moor2006nature,bostrom}.

One frequently encountered difficulty with value alignment is explicitly defining what constitutes a \textit{value}~\cite{soares2015value}.
Past approaches to AI value alignment have leveraged learning from observations or other forms of imitation learning~\cite{stadie2017third,Wulfmeier2019EfficientSF,ho2016generative}, the idea being that one can circumvent the requirement of value knowledge by learning to imitate human behavior instead. 
Learning knowledge from demonstrations which generalize beyond the context of the observation is difficult. Collecting sufficient demonstrations can be time consuming. Humans, too, are not necessarily able to comprehensively define a full set of principles or values even if asked to provide these examples.
%

Recent work has shown that stories are a promising potential source of value information.
In this work, authors use stories to learn a strong prior over behaviors. 
The issue with this work is that it was limited to a coarse view of value, opting to describe behavior as either being normative, aligning with expected social norms, or non-normative, deviating from expected social norms. 
Norms and the assessment of normative behavior can rarely be so neatly categorized into positive and negative valences in every context. 
In addition, it may be difficult to explain or remedy incorrect classification of normative behavior if systems lack an understanding of the specific \textit{normative principle} that is being violated. 
As such, additional work is needed to identify specific principles inherent in text-based descriptions and examples of normative behavior to better inform agents and the humans that work with these normative priors.

In this work, we seek to develop systems that have a more nuanced understanding of descriptions of human behavior with respect to \textit{normative principles}. 
We define a normative principles as specific behavior tenets that guide social normative behavior. An example of a potential normative principle might be \textit{Be polite to others.}
To facilitate this work, we augment a dataset of the children's comic strip, Goofus and Gallant, to contain detailed information about the principles being described in each frame. 
We then train various machine learning models with the aim of predicting the principle being either upheld or rejected based on images and text descriptions of the comics. 
We evaluate our work by comparing the performance of our trained models against humans who are tasked with performing a similar task.

\section{Related Work}

Value alignment encompasses the effort to have agents behave in accordance with human values without explicit control of a system's actions~\cite{bicchieri2005grammar,soares2014aligning,russell2015research,taylor2016alignment,arnold2017value,abel2016reinforcement}. These values - in and of themselves - are difficult to delineate. It can be confounding to attempt to discuss and define what specifically constitute human values. 

Various methods exist that attempt to instill moral reasoning or decision making within agents. Many of these methods, such as (cooperative) inverse reinforcement learning~\cite{ng2000algorithms,hadfield2016cooperative}, learning through imitation~\cite{ho2016generative}, and learning via expert demonstration~\cite{schaal1997learning,ho2016showing}, require substantial human input to generate training data. Furthermore, because these approaches train on limited and costly input, a strong prior distribution is necessary as is the case in language transfer with sparse training data demonstrated by Zoph et al~\cite{zoph2016transfer}.

Another approach that trains on natural language stories rather than demonstrated actions is known as Learning from Stories~\cite{riedl2016using,harrison2016learning}. This work proved successful at demonstrating the capability for agents to leverage reward signals from structures derived from narratives and perform more normatively than models which did not have similar priming. 


Other work attempts to control the output of generative models; the language model GPT was shown to be capable of generating sentences matching sentiment of training data~\cite{ziegler2019}. Though distinct from value alignment, this work demonstrates training models from data instilled with human characteristics and mannerisms.

One of the recent efforts to value alignment is Delphi~\cite{jiang2021delphi}, a model trained on unified moral dataset that can make moral judgments on real-world situations.
Despite being able to make moral judgement, Delphi does not give any notion of social norms or principles based on what it does make the judgement. 
Moreover, the dataset they have created to train the model were crowdsourced and from online platform i.e subreddit groups which consist of many inappropriate and societal biased set of examples. Therefore Delphi often makes improper and biased moral decisions. Another recent task on moral reasoning is Moral Stories~\cite{Denis2020Moral} which attempts to evaluate the capabilities of natural language generative model on making moral reasoning on social scenarios. The norms in Moral Stories are freeform sentences and very specific to situations (e.g. "It is wrong to disparage a classmate" or "You should return a jar of sugar once borrowed". This dataset was collected after our analysis and completion of this research. In contrast, the principles in our dataset are more general, broader and binned into a finite number of classes which is a more tractable starting point to train a value aligned agent. 




\section{Dataset}
In order to create our models of normative principle understanding, we must first find a suitable dataset for this task. 
To our knowledge, there is no pre-existing dataset that contains naturally occurring stories that are annotated with knowledge of normative principles. 
Thus, one of our primary contributions in this work is the curation of such a dataset. 
To construct this dataset, we use crowdsourcing to extend a previously collected dataset that was used to train binary classification models on the children's comic strcip, Goofus \& Gallant. 

This comic strip has been published in the U.S. children’s magazine, {\em Highlights}, since 1940 as a means to teach children socially acceptable behavior. It features two main characters Goofus and Gallant; portraying them in everyday common situations. The comic consistently portrays a pair of situations: both a proper way and an improper way to navigate typical situations that young adults may encounter. Thus Goofus \& Gallant comic strip is a natural corpus to categorize an action as normative or non-normative. But for the purpose of our work, the dataset as it exists is not sufficient. There is no specific identifying information that expands on what expectations the children may or may not be adhering to. We need to know which social norms or principles are violated or complied with by these actions.

Thus we use the crowdsourcing platform, {\em Prolific}, in order to enrich this dataset. For each image-action text pair of the comic, we collect a detailed text description of the scene in the image, refine these descriptions with additional annotations, and separately prompt other annotators to provide - in freeform text - a description of the social norm that is violated or upheld by the character in the comic. A comprehensive discussion of the data collection process is presented in the following section.

\begin{figure}[t]
\centering{
\includegraphics[scale=0.45]{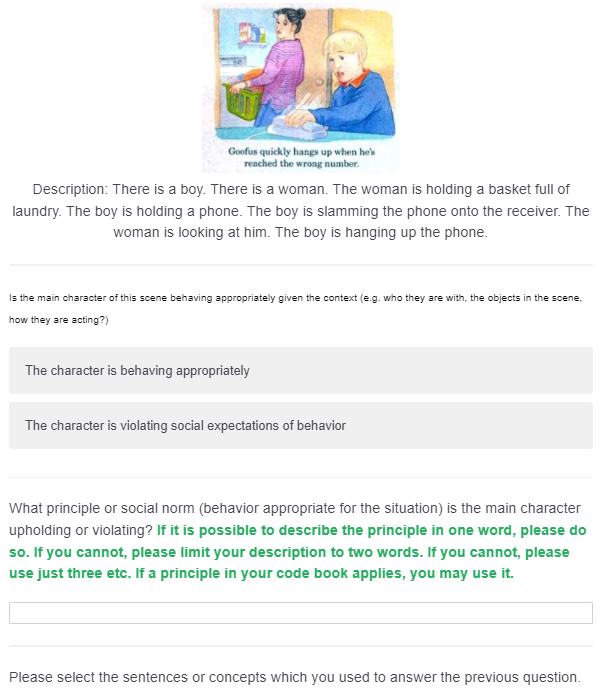}} 
\caption{Dataset collection prompt/survey}
\label{fig:dataworks_survey}
\end{figure}

\begin{figure}[t]
\centering{
\includegraphics[scale=0.4]{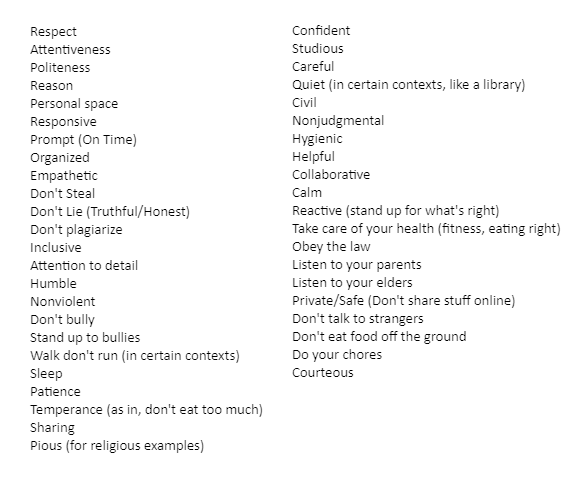}} 
\caption{Exemplar Principles List as provided in the prompt to crowd workers}
\label{fig:principleslist}
\end{figure}

\begin{figure}[t]
\centering{
\includegraphics[scale=0.45]{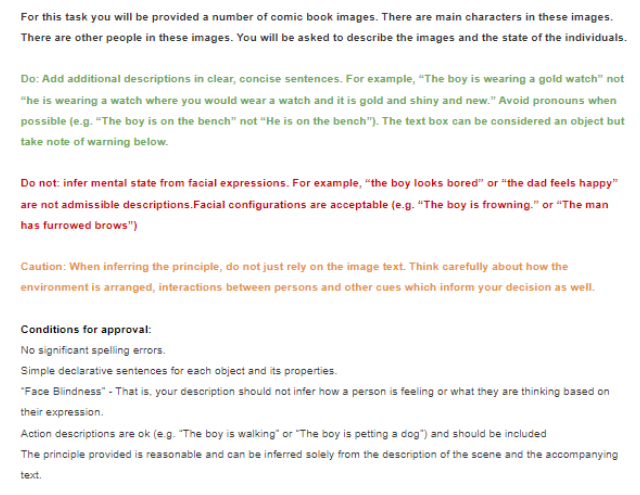}} 
\caption{Instructions given for the scene description task}
\label{fig:scene_description_instruction}
\end{figure}

\begin{figure}[t]
\centering{
\includegraphics[scale=0.45]{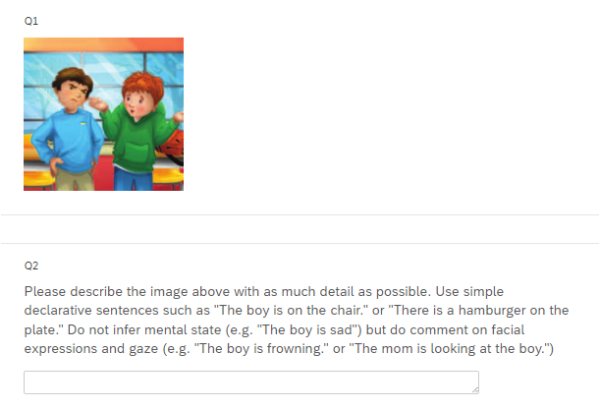}} 
\caption{Prompt and exemplar for scene description task survey}
\label{fig:scene_description_example}
\end{figure}

\begin{figure}[t]
\centering{
\includegraphics[scale=0.45]{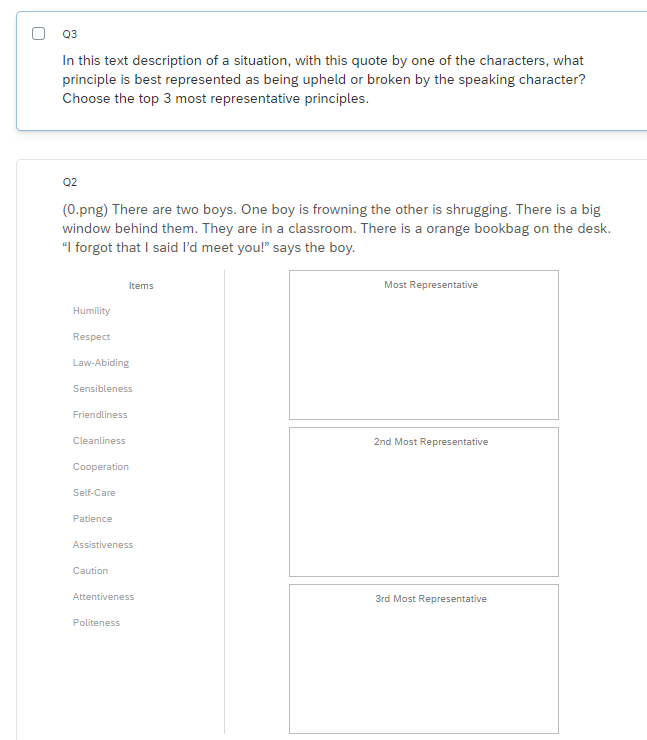}} 
\caption{Prompt and exemplar for "pick-and-rank 3" for 13 classes}
\label{fig:pick_rank_3}
\end{figure}
\subsection{Data Collection}
Recall, the goal of this data collection is to use human annotators to construct a dataset of Goofus \& Gallant comics that are annotated with normative principle information. 
To accomplish this, crowdworkers were recruited using a service in the [redacted] metro area. 
The crowdworkers were given a number of tasks, with no individual worker participating in more than one phase of the data collection. The tasks were as follows: 1) validate the original binary class in the Goofus and Gallant comic (i.e. - is this socially acceptable or unacceptable) (see Figure~\ref{fig:dataworks_survey}), 2) provide a description - in short, declarative sentences - of the comic image contents - but in a way that excludes assumptions about the theory of mind of the characters in the comics (e.g. not "The boy seems frustrated" only objectively observable information in the image) (see Figure~\ref{fig:scene_description_instruction}), 3) review in a secondary evaluation these descriptions written by other crowdworkers and remove incorrect observations or add missing descriptions and, finally, 4) Use the descriptions of the scene and just the image, so as to not know whether the comic depicts Goofus or Gallant, to describe which "social principle" is upheld or violated with light prompting of what a "social principle" is. Crowdworkers received images in tandem with the scene description and primary entity quote. In the comics, often the quote indicates Goofus is talking or Gallant is talking. The phrases were generalize to remove the identifying character (e.g. "'I'm bored,' says Goofus" becomes "I'm bored.)

For each of the tasks described, a workbook or template manual was provided which gave example responses, clarified terminology in the online survey they were instructed to take and explained the purpose of the experiment. The most relevant and "leading" component of these manuals or pamphlets was a list of exemplar "social principles" which were crowdsourced from our team (Figure~\ref{fig:principleslist}). The participants were never instructed to constrain their responses to this list in particular. Their responses and the principles they provided were always collected in freeform text.

We received 900 annotations from this data collection process. We remove annotations where annotators could not reach a consensus with the original Goofus \& Gallant tags from the comics (that is- annotators were not provided with text indicating the speaking character or who was being portrayed. 
The resulting dataset contains 772 comics annotated with normative principles. 
We, then, bin the freeform responses into 16 classes. These are then further reduced to 13 by eliminating the lowest frequency bins.
Authors of this paper independently grouped responses together into categories. This process was repeated until consensus was reached. 

\begin{figure*}[t]
     \centering
     \begin{subfigure}[b]{0.67\textwidth}
         \centering
         \includegraphics[width=\textwidth]{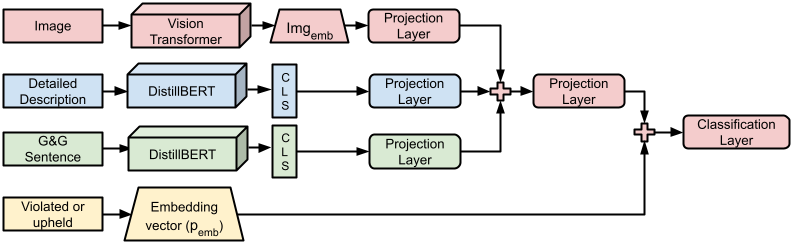}
         \caption{Model architecture for image and text inputs}
         \label{fig:img-text}
     \end{subfigure}
     \hfill
     \begin{subfigure}[b]{0.67\textwidth}
         \centering
         \includegraphics[width=\textwidth]{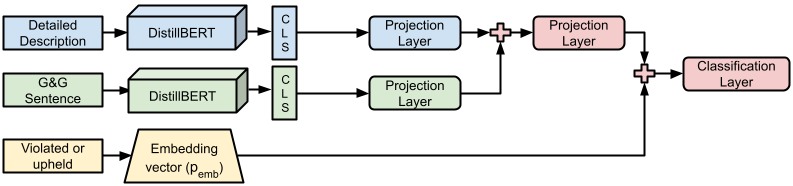}
         \caption{Model architecture for text only inputs}
         \label{fig:text only}
     \end{subfigure}
        \caption{Model architectures}
        \label{fig:model architecture}
\end{figure*}

\section{Methods}
The primary contribution of this paper is to explore how machine learning models can be used to learn normative principles on a naturally occurring story dataset. 

To do this, we developed two machine learning models that predict the normative principle involved in a Goofus \& Gallant comic.
In the first model, we classify principles using both image and text inputs and in the second, we inject only the text inputs into the model to investigate how influential/helpful visual context is for classifying principles. In both cases, we make use of proven transformer models - Vision Transformer~\cite{vit_alexey} and DistillBERT~\cite{distillBert} - as the basis for the network architecture. Each architecture is shown in Figure~\ref{fig:model architecture}. We discuss in detail each of the model’s architecture in following sections. 

\subsection{Image-Text Model}
For each comic image in the dataset, we now have the following: 1) text that comes from the original comic strip, 2) a detailed description of the image in simple declarative text as provided by annotators, and 3) the principle for the corresponding scenario in the image. For the first model, we pass both image and text information into the network. 
To give the network additional context, we also provide information on whether the principle in question is being violated or upheld int he form of a simplified binary vector. This corresponds to whether the original comic depicted Goofus (violated) or Gallant (upheld). 

In this work we use the pre-trained Vision Transformer~\cite{vit_alexey} to create the feature vector of the images. To embed text inputs - the detailed image description and comic sentence - we use a pre-trained DistillBERT~\cite{distillBert} model. 
We utilize the hidden representation of the special classification (CLS) token of DistillBERT as the embedding vectors. 
An identical layer is added on top of each pre-trained model which we refer to as the projection layer.
It consists of two linear layers followed by activation , dropout and layer normalization after each layer. 
All the embedding vectors are passed through these projection layers. The resultant vectors from the three projection layers are concatenated and passed through to another projection layer. 
The output vector of this layer is concatenated with the embedding vector $p_{emb}$. $p_{emb}$ represents the information, if the principle is violated or upheld in the current comic image. Finally this output vector goes into the classification layer, which is comprised of a linear and softmax layer. Figure \ref{fig:img-text} shows the overview of the network architecture of this model.

\subsection{Text Only Model}
The detailed description of an image in our dataset contains comprehensive descriptions of the scene and the state of the individual in the image. With this second model, we omit the image and only give the network the detailed scene description, the original comic text, and the vector indicating whether the normative principle is violated or upheld as inputs. With this model, we want to investigate the affect that the image has on predictive performance. 
This model is similar to the model described previously, except that elements related to learning image features have been removed. The overview of the network architecture is shown in Figure~\ref{fig:text only}.

\section{Experiments}
To evaluate the quality of our systems, we perform two sets of experiments: an automated evaluation and a human subjects evaluation. 
Each experiment is performed using both of our trained models. 
Each of these experiments will be discussed in greater detail below. 

\begin{figure}
\centering{
\includegraphics[scale=0.4]{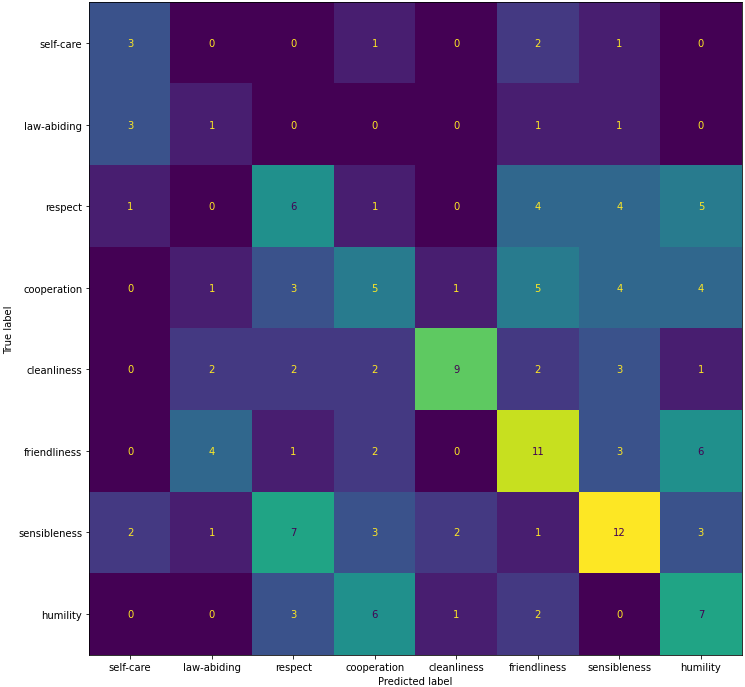}} 
\caption{Confusion matrix of the test data}
\label{fig:confusion_matrix}
\end{figure}

\begin{table*}
\centering
\footnotesize
\caption{Class Distribution and Test Accuracy for both Image-Text and Text-Only model with 13 Principles Dataset}
\label{table:results_13}
\begin{tabular}{ |p{2cm}||p{1.6cm}|p{1.6cm}|p{1.6cm}| p{1.6cm}| p{1.6cm}| p{1.6cm}| }
\hline
Class & Number of Data Points (Train) & Number of Data Points (Test) & Accuracy (image+text) & Accuracy (Text Only - top 1)  & Accuracy (Text Only - top 2) & Accuracy (Text Only - top 3)\\
\hline
\hline
Humility
&35&11&$27.27$&$36.36$&$63.63$&$72.73$\\
Respect
&85&21&$23.81$&$9.52$&$38.1$&$42.86$\\
Law-abiding
&32&6&$\textbf{0}$&$\textbf{16.67}$&$33.3$&$50.0$\\
Sensibleness
&11&2&$\textbf{0}$&$\textbf{0}$&$0$&$0.0$\\
Friendliness
&103&27&$37.04$&$40.74$&$48.15$&$55.56$\\
Cleanliness
&64&21&$\textbf{47.62}$&$\textbf{52.38}$&$\textbf{66.67}$&$\textbf{66.67}$\\
Cooperation
&49&16&$12.5$&$18.75$&$25.0$&$31.25$\\
Self-care
&29&7&$\textbf{0.0}$&$\textbf{14.29}$&$\textbf{28.57}$&$\textbf{28.57}$\\
Caution
&27&10&$\textbf{50.0}$&$\textbf{70.0}$&$\textbf{80.0}$&$\textbf{80.0}$\\
Patience
&34&4&$25.0$&$25.0$&$50.0$&$50.0$\\
Assistiveness
&35&7&$28.57$&$57.14$&$85.71$&$85.71$\\
Politeness
&53&8&$12.5$&$25.0$&$37.5$&$37.5$\\
Attentiveness
&60&15&$20.0$&$20.0$&$40.0$&$46.67$\\
\hline
\hline
\textbf{Totals/Averages}
&\textbf{617}&\textbf{155}&$\textbf{27.1}$&$\textbf{32.26}$&$\textbf{48.39}$&$\textbf{52.9}$\\
\hline
\end{tabular}
\end{table*}

\begin{table*}
\centering
\footnotesize
\caption{Class Distribution and Test Accuracy for both Image-Text and Text-Only model with 8 Principles Dataset}
\label{table:results_8}
\begin{tabular}{ |p{2cm}||p{1.6cm}|p{1.6cm}|p{1.6cm}| p{1.6cm}| p{1.6cm}|  p{1.6cm}|}
\hline
Class & Number of Data Points (Train) & Number of Data Points (Test) & Accuracy (image-text)& Accuracy (Text Only - top 1) & Accuracy (Text Only - Top 2) & Accuracy (Text Only - Top 3)\\
\hline
\hline
Humility
&$88$&$19$&$21.05$&$36.84$&$52.63$&$57.89$\\
Respect
&$85$&$21$&$28.57$&$28.57$&$38.1$&$52.38$\\
Law-abiding
&$32$&$6$&$16.67$&$16.67$&$33.3$&$66.67$\\
Sensibleness
&$132$&$31$&$\textbf{32.26}$&$\textbf{38.71}$&$\textbf{41.94}$&$\textbf{48.39}$\\
Friendliness
&$103$&$27$&$44.44$&$40.74$&$48.15$&$66.67$\\
Cleanliness
&$64$&$21$&$57.14$&$42.86$&$52.38$&$71.43$\\
Cooperation
&$84$&$23$&$26.09$&$21.74$&$65.22$&$78.26$\\
Self-care
&$29$&$7$&$\textbf{14.29}$&$\textbf{42.86}$&$\textbf{42.86}$&$\textbf{57.14}$\\
\hline
\hline
\textbf{Totals/Averages}
&$\textbf{617}$&$\textbf{155}$&$\textbf{33.55}$&$\textbf{34.84}$&$\textbf{48.39}$&$\textbf{61.94}$\\
\hline
\end{tabular}
\end{table*}

\begin{table*}
\centering
\footnotesize
\caption{Human classification (N=25) distribution and accuracy (Scene Description + Quote, No Image)}
\label{table:resultshumans}
\begin{tabular}{ |p{2cm}||p{1.6cm}|p{1.6cm}|p{1.6cm}| p{1.6cm}| p{1.6cm}| p{1.6cm}| p{1.6cm}|  }
\hline
Class & Accuracy (13 classes) & Accuracy (13-top2) & Accuracy (13-top3) & Accuracy (8 classes) & Accuracy (8-top 2) & Accuracy (8-top 3)\\
\hline
\hline
Humility
&$0\%$&$4\%$&$16\%$&$12\%$&$16\%$&$24\%$\\
Respect
&$16\%$&$28\%$&$40\%$&$28\%$&$60\%$&$80\%$\\
Law-abiding
&$4\%$&$8\%$&$32\%$&$28\%$&$36\%$&$48\%$\\
Sensibleness
&$\textbf{8\%}$&$\textbf{12\%}$&$\textbf{20\%}$&$4\%$&$16\%$&$28\%$\\
Friendliness
&$36\%$&$52\%$&$\textbf{68\%}$&$56\%$&$92\%$&$\textbf{96\%}$\\
Cleanliness
&$0\%$&$4\%$&$4\%$&$0\%$&$12\%$&$12\%$\\
Cooperation
&$16\%$&$24\%$&$52\%$&$48\%$&$60\%$&$64\%$\\
Self-care
&$0\%$&$8\%$&$12\%$&$12\%$&$28\%$&$40\%$\\
Caution
&$32\%$&$56\%$&$\textbf{64\%}$&$-$&$-$&$-$\\
Patience
&$36\%$&$48\%$&$60\%$&$-$&$-$&$-$\\
Assistiveness
&$4\%$&$16\%$&$20\%$&$-$&$-$&$-$\\
Politeness
&$28\%$&$48\%$&$56\%$&$-$&$-$&$-$\\
Attentiveness
&$32\%$&$36\%$&$36\%$&$-$&$-$&$-$\\
\hline
\hline
\textbf{Avg Accuracy}
&$\textbf{13.923\%}$&$\textbf{26.461\%}$&$\textbf{36.923\%}$&$\textbf{23.500\%}$&$\textbf{40.000\%}$&$\textbf{49.000\%}$\\
\hline
\end{tabular}
\end{table*}

\subsection{Automatic Evaluation Protocol}
The first set of experiments involves evaluating our methods using automatic performance metrics. 
We evaluate our models by holding out 20\% as a test set.
We use \textit{top1, top2}, and \textit{top3} accuracy for this evaluation. These metrics describe the percentage of correct predictions that appear in the top 1, 2, and 3 responses in terms of softmax probability, respectively. 

For this experiment, we used two different versions of our dataset for training. 
The first is the one described above containing 13 normative principles for classification. 
We also used a second version of this dataset that only contained 8 normative principles for classification. 
We have created this dataset by merging some of the principles from 13 principles set. For merging, we have mostly considered the principles which have relatively lower number of data points and then merged them based on their context similarity. For instance, Cooperation and assisting have similar meaning. Similarly, being patient is also a showcasing of sensibleness, hence these concepts may indicate similar situations.
By merging principles that are related, we aim to investigate how well our methods perform on a smaller set of principles that are less likely to be correlated. 

\subsection{Human Subjects Evaluation Protocol}
We also perform a human subjects experiment comparing the performance of our models to the performance of humans on the task of predicting normative principles based solely on text representations of scenes. The decision to use only text was grounded in our belief that the task is sufficiently difficult for humans even with slightly modified text from the comic still augmented with more descriptive text containing what existed in the original image. This better mirrors original attempts to use text-only transformers on the classification task.
In this experiment, crowd workers from Prolific were presented with text-only descriptions of Goofus and Gallant comics drawn paired with the original comic text. 
Comics were presented at random - one representative image with a groundtruth principle tag - from our dataset containing 13 normative principles.
Workers were tasked with selecting and ranking the top three principles from a list they were presented that described the comic in question.

Workers completed 5 ranking tasks at a time - randomly selected but evenly distributed among the 13 core examples. 
In total, across all images and participants, the pick-and-rank task was completed 625 times (Figure~\ref{fig:pick_rank_3} shows the data collection interface). Each of the 13 examples description-quote pairs received 25 rankings (that is, 65 participants chose their top 3 representative principles for the 5 images presented during their task set). A total of 25 rankings per principle were collected as a result.

This experiment was repeated using our downselected dataset that contained only 8 normative principles 
This experiment involved enough workers to achieve 25 rankings per principle as in our previous experiment.

Both experiments were evaluated using the \textit{top1, top2}, and \textit{top3} accuracies, as we did for our automatic evaluation. 
Here, these accuracy metrics describe how often human participants correctly identified the normative principle for a comic in their top 1, 2, and 3 responses respectively.

\section{Results}

\subsection{Automatic Evaluation}
The prediction results of the models for both 13 and 8 principles are shown in the Table~\ref{table:results_13} and~\ref{table:results_8} respectively. 
Both tables show the accuracy of our two models: 1) Image-Text model and 2) Text-Only model. From the table we can observe, though injecting visual information into the model improves the accuracy for some of the classes but the overall performance is decreased.
It indicates, the visual cues such as individual’s facial expression, surrounding objects contribute very nominal in predicting social principles.
Instead the textual description of the scene and action dominantly influence in understanding the principles. Thus in our later analysis we only refer to the result of the Text-Only model.

Accuracy for some of the classes are relatively lower for 13 principles set as they have smaller number of data points in the training set, for instance, “Sensibleness” (Table~\ref{table:results_13}). The accuracy of these classes increases significantly after downsizing the classes. From Table~\ref{table:results_8}, we can see that, the model's capability to predict "Self-care", "Sensibleness" improves considerably than the model trained with 13 principles. But it is worth to mention that, though downsizing the number of classes increases model's performance for some of the classes, the overall accuracy does not increase largely. 


\subsection{Human Subject Evaluation}
The results of the human subject evaluations on text-descriptions and quotes (with no image) can be seen in Table ~\ref{table:resultshumans}. The results for 13 principles are presented along side the results for 8 principles. A key observation we can make is how this accuracy shifts when the principle list is reduced. Ambiguous or more infrequent principles are absorbed and the annotators take less time on the task, with fewer principles to deliberate between. It is worth noting that only in the case of {\em Sensibleness} (aka "Sensibility") did accuracy decline. In all other cases, the further binning of principles from 13 to 8 greatly increased the capability of human annotators with respect to correctly identifying the principle.


\section{Discussion}
In this section, we discuss the results of both our automatic evaluation and our human subjects evaluation. 

\subsection{Automatic Evaluation}
The first thing to note about the results of our automatic evaluation is that the performance, overall, of each model is relatively poor. 
The top1 performance for both the text only and text+image models are below 50\%.
An interesting note is that the presence of the image did not improve prediction performance. 
In fact, the machine learning model that utilized image features did worse than the model that only had access to text features.
It is for this reason that we focused our evaluations on the text only model. One possible explanation is how the comic image stylization has changed over the decades - the dataset was already fairly sparse and so these differences likely had a significant impact.

In addition, we see that our model's overall average accuracy increases as we consider wider rangers for accuracy. 
This lends support to the notion that many of our principles may conceptually overlap with each other. 
The models struggle to differentiate between principles that may occur in similar situations (i.e. "Cooperation" and "Humility", can be seen from the confusion matrix shown in Figure~\ref{fig:confusion_matrix}), making it less likely that the correct answer appears as the top response, but more likely that it appears in the top 3 responses.
This idea is further supported by the overall increase in performance we see when moving to the downselected dataset. 
By merging certain principles together, we enable the machine learning model to better differentiate between principles, leading to overall better predictive accuracy.


\subsection{Human Subject Evaluation}
Similar to the difficulty large scale language models faced when provided text-only descriptions of the comics, human participants struggled to accurately identify principles when given the same prompt. It is intuitive that the accuracy improves when there are fewer principles to choose from. In many cases, it may not be unreasonable to apply multiple principles to a given situation or situational description or a quote from a peer. One interpretation for the ambiguity of the results may also be the nature of the original collection methods. The crowdworkers asked with attributing freeform principles to the comics likely have significantly diverse mental templates, expectations and memories of what "cooperation" may mean as opposed to "assistiveness" as one example. When asking another set of participants to select, even from a much reduced set of principles, this continues to be a problem. But we do see with principles like "friendliness" or "caution" - two which receive fairly high accuracy/consensus - that there are concepts, situational descriptions and prompts which more clearly represent a subset of the binned principles. Another explanation for "friendliness"'s high performance across both bins may be that it becomes the default principle participants choose when all others are confusing. Indeed, some principles may be pre-requirements to others. If a person is effective at the other principles, they are likely to be perceived as "friendly" in general.

Perhaps the most important thing to note is how our machine learning models performed with respect to the human rankings. 
If one looks at average accuracy, our models outperformed humans across all metrics on both the 13 principle dataset and the 8 principle dataset. 

\section{Conclusion}

It is important and urgent to have more perspectives on how best to align autonomous systems with human preferences, values and social norms. The task remains difficult despite new datasets or other methods which focus on debiasing or particular moral philosophies.  The principles which were collected can be applied to more than the western cultures specifically depicted in the dataset which was extended. It may be useful to investigate the mapping between descriptions of situations and principles that were collected. We also see that the task of identifying socially-normative principles is difficult for both human annotators as well as complex, state of the art language models and custom architectures. It is not unreasonable to assume additional context is needed; the theory of mind of entities being assessed, previous actions and future outcomes of similar social normative situations are part of ongoing work to probe even deeper into this problem. We hope this facilitates discussion as to how best to expand existing datasets to understand normative behavior in terms beyond simple "acceptability" and "non-acceptable" behaviors or those outside traditional Western moral frameworks.

\bibliography{acl_latex}


\end{document}